\documentclass[conference,a4paper]{IEEEtran}

\usepackage{cite}
\usepackage{amsmath,amssymb,amsfonts}
\usepackage{graphicx}
\usepackage{textcomp}
\usepackage{stmaryrd}
\usepackage{xcolor}
\usepackage{adjustbox}
\usepackage{algorithm}
\usepackage{algpseudocode}
\DeclareMathOperator*{\argmin}{\arg\!\min}
\def\BibTeX{{\rm B\kern-.05em{\sc i\kern-.025em b}\kern-.08em
    T\kern-.1667em\lower.7ex\hbox{E}\kern-.125emX}}

\def\eg{\emph{e.g}. }

\def\etal{\emph{et al}. } 

\begin{document}
\IEEEoverridecommandlockouts
\IEEEpubid{\makebox[\columnwidth]{979-8-3503-7903-7/24/\$31.00 ©2024 European Union \hfill} \hspace{\columnsep}\makebox[\columnwidth]{ }}
\title{LatentForensics: Towards frugal deepfake detection in the StyleGAN latent space}

\author{\IEEEauthorblockN{Matthieu Delmas}
\IEEEauthorblockA{
\textit{CentraleSup\'elec, IETR (UMR CNRS 6164)}\\
Rennes, France \\
matthieu.delmas@centralesupelec.fr}
\and
\IEEEauthorblockN{Renaud Seguier}
\IEEEauthorblockA{
\textit{CentraleSup\'elec, IETR (UMR CNRS 6164)}\\
Rennes, France \\
renaud.seguier@centralesupelec.fr}
}

\maketitle

\begin{abstract}
   Classification of forged videos has been a challenge for the past few years. Deepfake classifiers can now reliably predict whether or not video frames have been tampered with. However, their performance is tied to both the dataset used for training and the analyst's computational power. We propose a deepfake classifier that operates in the latent space of a state-of-the-art generative adversarial network (GAN) trained on high-quality face images. The proposed method leverages the structure of the latent space of StyleGAN to learn a lightweight binary classification model. 
   Experimental results on standard datasets reveal that the proposed approach outperforms other state-of-the-art deepfake classification methods, especially in contexts where the data available to train the models is rare, such as when a new manipulation method is introduced.
    To the best of our knowledge, this is the first study showing the interest of the latent space of StyleGAN for deepfake classification. Combined with other recent studies on the interpretation and manipulation of this latent space, we believe that the proposed approach can further help in developing frugal deepfake classification methods based on interpretable high-level properties of face images.
\end{abstract}

\begin{IEEEkeywords}
Deepfakes, Computer Vision, Latent Space, Frugality
\end{IEEEkeywords}

\section{Introduction}
Forgery of videos, the creation of so-called deepfakes, has been on the rise for the past few years. Although it yields multiple benefits in different domains (\eg special effects and data generation), its democratization comes with a few risks. It is now easier than ever for malevolent users to create fake media in order to discredit trusted sources, impersonate powerful political figures, or blackmail individuals.
Fortunately, recent work in media forensics have yielded numerous ways to discriminate deepfakes from genuine videos. Most of those methods rely on the use of Convolutional Neural Networks (CNNs)\cite{lecun1989backpropagation} which are tailored to detect specific weaknesses or artifacts left by the forgery process \cite{afchar2018mesonet} \cite{chollet2017xception}. However, this means that as the forgery process improves, the cost of training such discriminators will increase.
As sustainability becomes an ever-increasing priority, working towards building models which require less resources such as training examples, computing power and time -also known as frugal AI- is a growing subject in the deep learning community.

A forensic method is proposed here to classify forged face images from genuine ones. Its training requires considerably less resources than current state-of-the-art models while still remaining competitive or even achieving better performance.

\begin{figure*}[]
\centering
\includegraphics[trim={0.7cm 1cm 0.8cm  0cm},clip,scale=0.88]{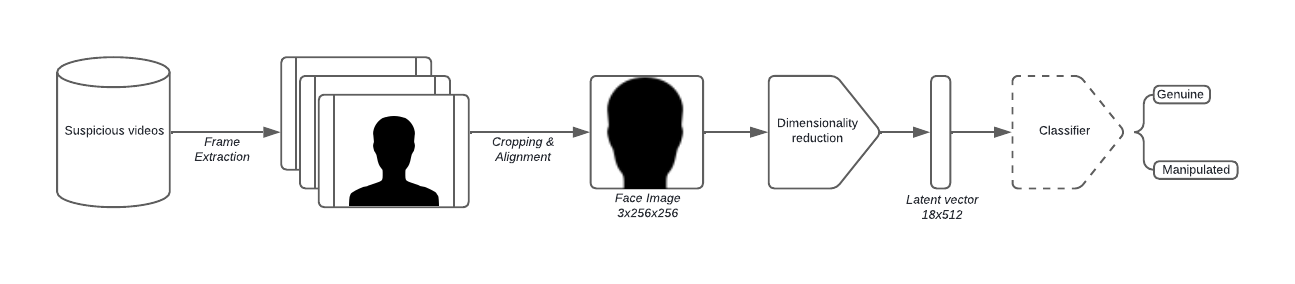}
\caption{The proposed pipeline, in dotted lines the trained model.}
\label{pipelinefig}
\end{figure*}

Recently, artificial image generation through Generative Adversarial Nets (GANs), notably of faces, has seen great progress with models such as StyleGAN \cite{karras2019style} or CycleGAN \cite{zhu2017unpaired} in both the quality of results and the understanding of the generation process. It is proposed to capitalize on those improvements by projecting suspect image data in the latent space of StyleGAN before performing the classification.
The resulting pipeline, represented Figure \ref{pipelinefig}, is quite simple but its strength resides in its combined efficiency and how easy it is to set up, especially compared to other state-of-the-art models.

Our contributions are as follows:

\begin{itemize}
  \item A comparison  of dimensionality reduction methods, including StyleGAN pseudo-inversions in the context of deepfake classification.
  \item An easy-to-train and effective deepfake classifier which works on images inverted in the latent space of StyleGAN.
  \item A demonstration in the context in which the proposed method is best-suited : when only limited data or computing power is available.
\end{itemize}

With the goal of reducing the dimensonality of the data in mind, a first study is made to compare the latent space of StyleGAN to a Principal Component Analysis (PCA) on the DeepfakeDetectionChallenge (DFDC) preview dataset \cite{dolhansky2019deepfake}. Then, as there are numerous ways to obtain a StyleGAN latent code from an image, a benchmark of different StyleGAN inversion methods is made to find the most suitable in a deepfake classification context. Next, with the best possible tools, an end-to-end binary discrimination pipeline is set up and compared to state of the art models (MesoNet\cite{afchar2018mesonet}, XceptionNet\cite{chollet2017xception} and EfficientNet\cite{tan2019efficientnet}) on the CelebDF v2 \cite{Celeb_DF_cvpr20} database. A particular importance is given to computing and data requirements to train the models.
Finally, the approach is further pursued by studying how the proposed approach performs on small databases.
\section{Related Work}
There are two main trains of thought when it comes to discerning a deepfake from a genuine video: artefact detection and subject identification. In the former method, models trained on images spot where manipulations have been made, these include MesoNet \cite{afchar2018mesonet} or XCeptionNet \cite{chollet2017xception}. Such models rely on inconsistencies in image patches, \textit{i.e.} the presence of visual artifacts left by the forgery process. On the other hand, identification methods train a model to try to recognize if the portrayed person is the expected subject or an impersonator, instead of trying to spot if a certain frame has been tampered with.

\subsection{Detection of Image Artifacts}
CNNs have been achieving top performance in analyzing video frames for deepfake classification. MesoNet \cite{afchar2018mesonet}, for instance, is a lightweight convolutional network which relies on the density of details in image patches to detect whether or not an image have been tampered with.
Such networks have been state of the art for a while, as highlighted by Rossler \etal in their FaceForensics++ (FF++) study \cite{rosslerfaceforensics}.
More recently, the Deep Fake Detection Challenge \cite{dolhansky2020deepfake} (DFDC) allowed multiple teams to compete in providing the best detection methods. 
Models based on FAb-NEt \cite{wiles2018self} came on top and achieved consistent performance.
These neural networks are arguably one of the best  solution to the deepfake detection problem, if one has enough computing power, training data, and if the deepfake manipulation method is known in advance. However, even though such methods hold great performance on given datasets and on raw images, success rates tend to drop when they are faced with unseen, compressed or altered data, as highlighted by the FF++ \cite{rosslerfaceforensics} and DFDC \cite{dolhansky2020deepfake} studies. As such, basing the detection process on raw image analysis might not keep holding up in the future, in the adverserial race between deepfake detectors and creators.
Furthermore and unfortunately, efforts have seldom been made to reduce the computational costs in training high-quality deepfake detectors.

\subsection{Semantics-based Classification}
To face this growing problem, methods have been put together to circumvent the need of having to rely on ever-subtler image artifacts, and to make the decision at a higher, more semantically robust level.

Agarwal \etal \cite{agarwal2020detecting} and Cozzolino \etal \cite{cozzolino2021id} combine neural networks and statistical methods to extract a vector representing the identity of the speaker, and then compare it to one from genuine footage. However, while such methods tend to overfit less on particular manipulations, they need access to supplementary information compared to traditional CNN-based methods (\eg{}  footage which is known to be genuine, or temporal data). 

In the present paper, the proposed method takes inspiration from these approaches. Instead of searching for manipulation artifacts, a good understanding of the deepfake creation process may allow for more efficient and robust detection protocols.

\section{Method}
This section first explains why it could be desirable to perform deepfake classification on a low dimensionality space, and then describes how such a dimensionality reduction might be performed, with an emphasis on StyleGAN inversions.
\subsection{Overview}

Basing the discrimination on a low-dimensional space in which to represent the suspicious frame may ease deepfake classification.
Intuitively, after reducing the dimension of the input, details which could mislead the model (\eg background, pose, lighting, compression noise) can be erased. The model could then focus on the most important information (\eg ``Are the subject's expressions coherent ?"). Furthermore, while CNNs achieve great performance in computer vision tasks, they require a lot of training data. A low-dimensionality model could be lighter, cheaper and easier-to-train if not based on CNN architectures. The curse of dimensionality makes the amount of data points required in finding a discrimination frontier exponentially more complex as the number of dimensions increases. As such, one could reasonalbly think that if the deepfake data is projected in a space of lower dimensionality, fewer training points would be required to build a robust model.

A good projection space in this case is an efficient representation of the faces, which would make the distinction between the distribution of real faces and manipulated ones easy.
The resulting decision barrier may even be more robust to slight changes in the data distribution (from compression, different context, etc.). However we need the projector to keep as much statistics relevant to the deepfake detection process (intuitively information such as the subject's identity and expressions for instance) as possible.

Formally, the projector is a function $P:\mathbb{R}^{l\times h\times w} \to \mathbb{R}^{d'}$ , with $d = l\times h\times w$ the dimension of the original high-dimensional image and $d' \ll d$ the dimension of the lower-dimensional latent representation.

\subsection{Dimensionality Reduction Algorithms}

There are many ways to reduce data dimensionality: Principal Component Analysis (PCA) \cite{pearson1901liii} can compress data while keeping as much variance as possible, Autoencoders \cite{kramer1991nonlinear} and Variational Autoencoders (VAEs)  \cite{kingma2013auto} are a (regularised) non-linear extension of this technique. Only the PCA method will be considered in the context of the proposed approach.

\begin{figure}[]
\centering
\includegraphics[trim={7cm 0cm 7cm  0cm},clip,scale=0.32]{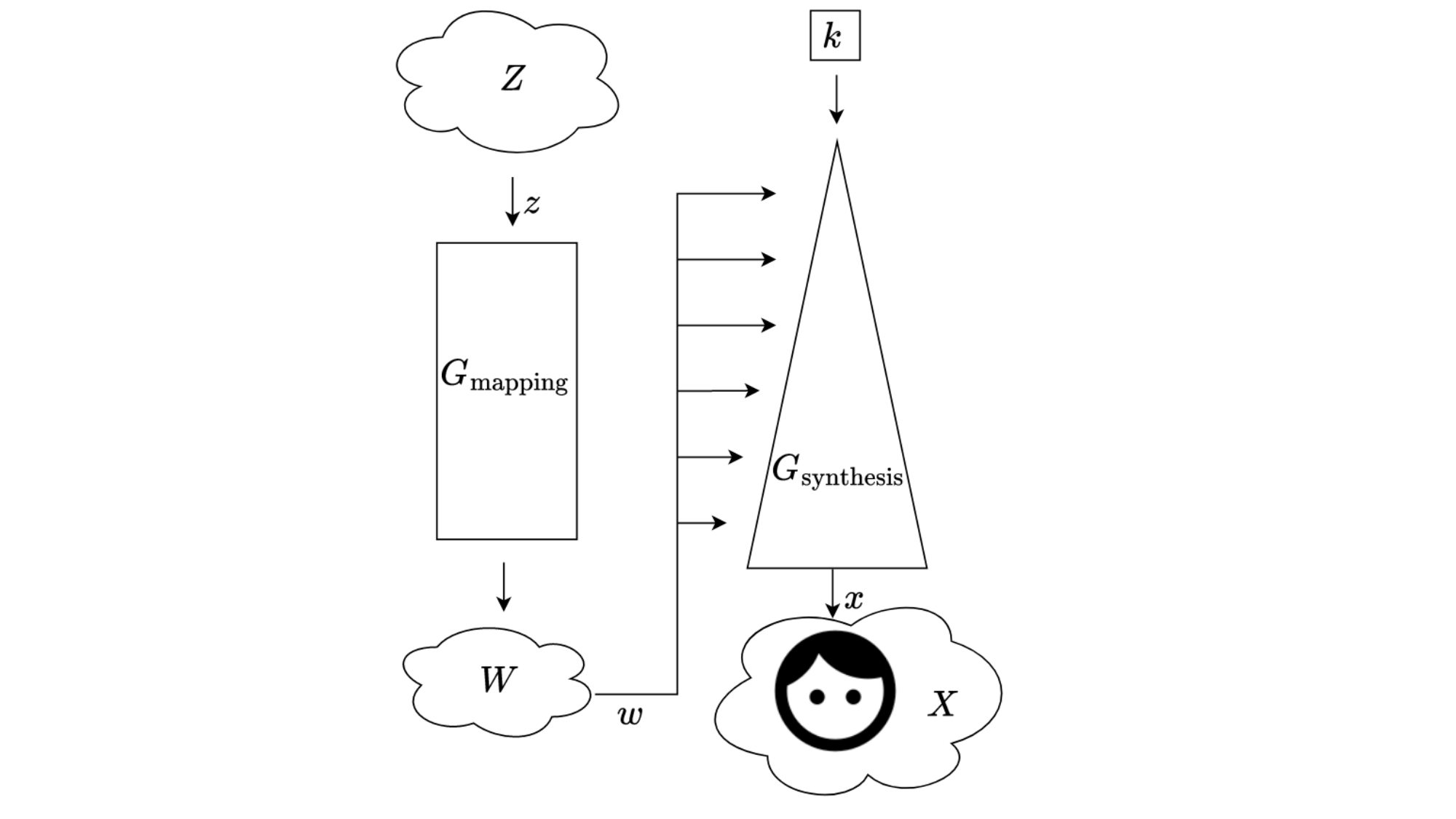}
\caption{The StyleGAN architecture summarised, $k$ is a constant learned during training.}
\label{stylegan}
\end{figure}

For images of faces in particular, the generative network StyleGAN \cite{karras2019style}, on top of producing images of excellent quality, has the advantage of having a unique architecture which generates data via an intermediate latent space. The structure of the generator network is summarized Figure \ref{stylegan}. A normally sampled variable $z,  z \sim \mathcal{N}(0,1)^{512}$ is first transformed to a latent code $w$ through a mapping network $G_{\text{mapping}}(z) = w$. In turn, this variable is used at different stages of the proper image generation, done by another neural network $G_{\text{synthesis}}$.

The Latent space $W$ defined as $w \in W, G_{\text{mapping}}(z) = w$ fits the criteria of a good support for the projected data: we already know there is a function from it to the face image space, and that it is of low dimensionality. In it, significant semantic directions have already been found \cite{shen2020interpreting}, \cite{tewari2020pie}, \cite{harkonen2020ganspace}, which suggest its structure is ideal to study or perform facial modifications which are easily interpretable by humans.

Multiple methods have been proposed by the community in the past years to obtain a suitable StyleGAN latent code from an image (real or generated), each with their own advantages and drawbacks.

To invert suspicious frames in the latent space of StyleGAN, the first proposed methods went through an optimization process, where the latent code is progressively modified so that the output of StyleGAN matches the desired image. Many solutions exist for this task, such as proposed originally by \cite{karras2019style} but also by  \cite{abdal2019image2stylegan}, \cite{zhu2020domain} and \cite{xia2022gan}. Most of these solutions are based on gradient descent-algorithms to minimize a loss such as $L(w) = L_{p}(G_{\text{synthesis}}(w), x) + \alpha ||G_{\text{synthesis}}(w) - x||_2^2$, with $L_{p}$ a neural distance such as LPIPS \cite{zhang2018perceptual}. 

The drawback of those methods is that unfortunately, the gradient descent optimization may take a few thousands steps to converge, along with problems that typically arise in optimization  (\eg presence of local minima, high instability). More recently, researchers developed encoder-based model which are trained to output the latent code from an image, either before performing later refinements, such as IDInvert \cite{zhu2020domain}, and E2Style \cite{wei2022e2style} or directly, such as by Psp \cite{richardson2021encoding}. These models are described in more details in section \ref{dimrecsubsection}.

The proposed method consists of using these StyleGAN pseudo-inversion models to obtain low-dimensionality latent codes, on which deepfakes classifiers will be trained. A study of different latent spaces, inversion methods and deepfake classifiers is presented in the following section.

\section{Experiments}
A few experiments are described here to discuss the usefulness of using the latent space of StyleGAN to perform deepfake discrimination. Section \ref{protocolsubsection} describes the experimental setup put in place. In section \ref{dimrecsubsection}, it is first shown how this latent space is better suited than a standard Principal Components Analysis as a dimensionality reduction method. A comparison of different StyleGAN inversion methods is then proposed, before looking at how one could further reduce the data dimensionality using particular channels in the StyleGAN latent space. In section \ref{fullpipesubsection}, the full pipeline, as described Figure \ref{pipelinefig}, is tested against state-of-the-art deepfake detection methods, at first on the CelebDFv2 dataset, and finally on smaller databases.

\subsection{Models used and Data Preparation} \label{protocolsubsection}
The presented experiments follow the pipeline shown in Figure \ref{pipelinefig} : after a temporal sampling of suspicious frames, images of faces are extracted, cropped and centered before having their dimensionality reduced. We then train a classifier to discriminate between the latent vectors originating from deepfakes and genuine videos.
As the main focus of the study is the influence of the dimensionality reduction, we kept the training parameters of the classifiers simple. Unless otherwise noted,  classification models used after the dimensionality reduction are neural networks composed of five fully-connected layers with ReLU activation functions, trained to minimize the Binary Cross Entropy Loss at a learning rate of $0.005$. The network has 2048 neurons in its input layer, 512 neurons in all three middle layers, and a single neuron in its last layer.
In the pipeline, only the classifier (in dotted line in Figure \ref{pipelinefig} is trained by the experimenters, all the other parts are used as is.

To conduct these experiments, three datasets have been chosen : the Deepfake Detection Challenge (DFDC) preview dataset \cite{dolhansky2019deepfake}, as well as CelebDF version 1 and 2 \cite{Celeb_DF_cvpr20}. The former two datasets are of smaller scale than CelebDF v2. They were used for the lighter experiments in section \ref{dimrecsubsection} for practical purposes, as well as in section \ref{fullpipesubsection} to especially study the performance of the proposed method on smaller datasets. The biggest and most challenging dataset, CelebDF v2, was used for the main comparison in section \ref{fullpipesubsection}.

For every dataset, about one in ten frames was uniformly sampled throughout the videos, this sampling was made to limit computing requirements, as well as to ensure a relatively diverse set of faces, not too correlated to one another.
Each frame was processed according to the Flickr-Faces-HQ (FFHQ) \cite{karras2019style} alignment protocol before having their dimensionality reduced, usually by inverting the images of the faces in the StyleGAN latent space. Training, validation and testing sets were separated at the video level, with 10\% of the dataset being reserved for testing.
Table \ref{datarepart} shows the number of examples in each dataset.

\begin{table}[]
\begin{center}
\begin{tabular}{l|c|c}
Dataset & \# Videos & \# Frames\\
\hline
CelebDF v1 & 1203  & 23 902\\
DFDC preview & 5214 &  178 668\\
CelebDF v2    & 6529 & 250 972\\

\end{tabular}

\end{center}
\caption{Description of the datasets used}
  \label{datarepart}
\end{table}

Aside from the Principal Components Analysis studied section \ref{dimrecsubsection}, every dimensionality reduction method was taken pre-trained on StyleGAN and was frozen since. This was a deliberate choice to keep the models agnostic relative to the deepfake detection method, and to ensure that the proposed method was still requiring few resources to train.

\subsection{Dimensionality Reduction Comparison} \label{dimrecsubsection}
A study is proposed here to confirm the efficacy of dimensionality reduction in the context of deepfake discrimination, and to benchmark the pseudo-inversion of StyleGAN against PCA.

To invert real world images of faces in the latent space of StyleGAN, the IDInvert \cite{zhu2020domain} method is first considered. It is  a well-known encoder-optimization hybrid approach which has been proposed as an early alternative to costly inversions based solely on optimization. 
An encoder network $E$ is trained to produce an initial latent code of size $14 \times 512$ with domain knowledge (in this case images of faces) to improve on the inversion quality. The latent code outputted by the encoder $w_0 = E(x)$ is then optimized further to improve the results. 

The network is trained to minimize a combination of distances calculated in a version of $W+$ space with 14 channels, the image space, and the feature space of the VGG network \cite{simonyan2015very} $F(x)$. With $\lambda_{\textit{$vgg$}},  \lambda_{\textit{$dom$}}$ two weights, the resulting problem to optimize is described equation \ref{IDeqn}.
    \begin{equation}
    \begin{split}
    w^* = \argmin_w (||x - G_{\text{synthesis}}(w)|| \\
    + \lambda_{\textit{vgg}}||F(x) - F(S(w))||_2 \\
+ \lambda_{\textit{dom}}||w - E(G_{\text{synthesis}}(w))||_2)
    \end{split}
    \label{IDeqn}
    \end{equation}
The IDInvert model was used pre-trained on the FFHQ dataset, for a fair comparison, an Incremental PCA model \cite{ross2008incremental}, implemented with the scikit-learn package\cite{scikit-learn} was trained on the same dataset. This PCA model kept the 512 dimensions which explained the most variance. This number was chosen as the original StyleGAN model used a latent space of same dimensionality.

To test if there is a dimensionality reduction method better suited for discriminating deepfakes, different instances of a Random Forest classifier \cite{breiman1996out} are trained on the codes outputted by each projector. Those classifiers were composed of 1500 estimators and were trained to optimize the Gini criterion. They were implemented using scikit-learn , with other parameters left to default. The results are shown in Table~\ref{resRFDim}.

\begin{table}[]
\begin{center}
\begin{tabular}{l|c}
Dimensionality Reduction & Accuracy\\
\hline
PCA & 0.78  \\
StyleGAN Inversion    & \textbf{0.88}  \\
\end{tabular}

\end{center}
\caption{Comparison of Random Forest Classifiers detection accuracy on low dimensionality data}
  \label{resRFDim}
\end{table}

To classify deepfakes, the StyleGAN inversion method obtains the best performance by a considerable margin (10 percentage points).

This analysis shows that a few thousand dimensions, instead of the hundreds of thousands usually used, can contain enough information to discriminate genuine images from deepfakes to a satisfactory degree, even by models as simple as Random Forest classifiers. This motivates the progression towards a more frugal approach to deepfake classification.

 A comparison of three of the most used StyleGAN inversion models (Psp\cite{richardson2021encoding}, E2Style\cite{wei2022e2style} and HFGI\cite{wang2022high}) on top of IDInvert, is now proposed here to determine which one is the most suitable for deepfake detection.

\begin{itemize}
               
    \item  Psp introduces eighteen independent modules ``map2style'', noted $\beta$, which transform the latent feature maps of a pre-trained ResNet $R$ into style vectors from $W+$ such as $\forall i \in \llbracket 1~;~ 18 \rrbracket , w_i = \beta_i(R(x))$.
  
    \item  E2Style encodes an image to a first latent code $w_0 = E_0(x)$, which can be iteratively refined by different encoders $w_t = E_t(x, G_{\text{synthesis}}(w_{t-1}))+w_{t-1}$. Each encoder $E_t$ are trained separately in order to have the advantages of the hybrid approach without having to go through an optimization process.
    \item  HFGI is one of the most recent approach. A first latent code is encoded $w_0=E_0(x), \hat{x}_0=G_{\text{synthesis}}(w_0)$, a second network $E_c$ then will transform the reconstruction error $\Delta = x-\hat{x}_0$ in another latent code. These two vectors are then merged by a module $D$ which produces a final StyleGAN latent code $\hat{x}= G_{\text{synthesis}}(D(w_0, E_c(\Delta)))$.

\end{itemize}
The aforementioned five layer fully-connected network is trained to discriminate between genuine and deepfake latent codes obtained from the different inversion methods.
 Both the binary discrimination accuracy of the trained models and the time it took to invert an image in the latent space are shown in table \ref{resTable}. The inversion speed has been measured on a NVIDIA QUADRO RTX 3000 graphics card.

\begin{table}[]
\begin{center}
\begin{tabular}{l|c|c|c|c|c}
Inversion & Accuracy (\%) & Throughput (s/frame)\\
\hline
IDInvert \cite{zhu2020domain} & 94.0 & 86.9 \\         
Psp  \cite{richardson2021encoding} & 96.31 & 0.0492 \\          
HFGI \cite{wang2022high}  & 96.43 & \textbf{0.0396} \\        
E2Style \cite{wei2022e2style} & \textbf{98.46} & 0.0618\\
\end{tabular}

\end{center}
\caption{Binary deepfake detection accuracy and speed depending on the inversion method.}
  \label{resTable}
\end{table}

It is becoming clearer why dimensionality reduction can be relevant in deepfake discrimination, while earlier, optimization-based inversion methods took too long to be viable, more recent models allow for quick and efficient dimensionality reduction (as much as 2 000 times faster inversion), while further improving the quality of the results, both visually, as shown in their respective papers, but also in deepfake discrimination.

While the current research in StyleGAN inversion usually trends towards producing latent codes in the W+ space (dimensionality of $18\times 512$), the original StyleGAN was supposed to work with latent codes of size 512 (the W space).
Furthermore, experiments in style mixing, namely the ones performed in the original paper \cite{karras2019style}, demonstrated how the different steps in the generation of the data (which corresponds to different channels in W+ space) influenced the final results in drastically different ways. Early modification of the latent code resulted in modification of low level facial features, such as age, gender or global shapes, while later modifications influenced finer details, such as skin tone and hair color, without modifying the global shape of the face.

\begin{figure}[]
\centering
\includegraphics[trim={0cm 0cm 0cm  0cm},clip,scale=0.32]{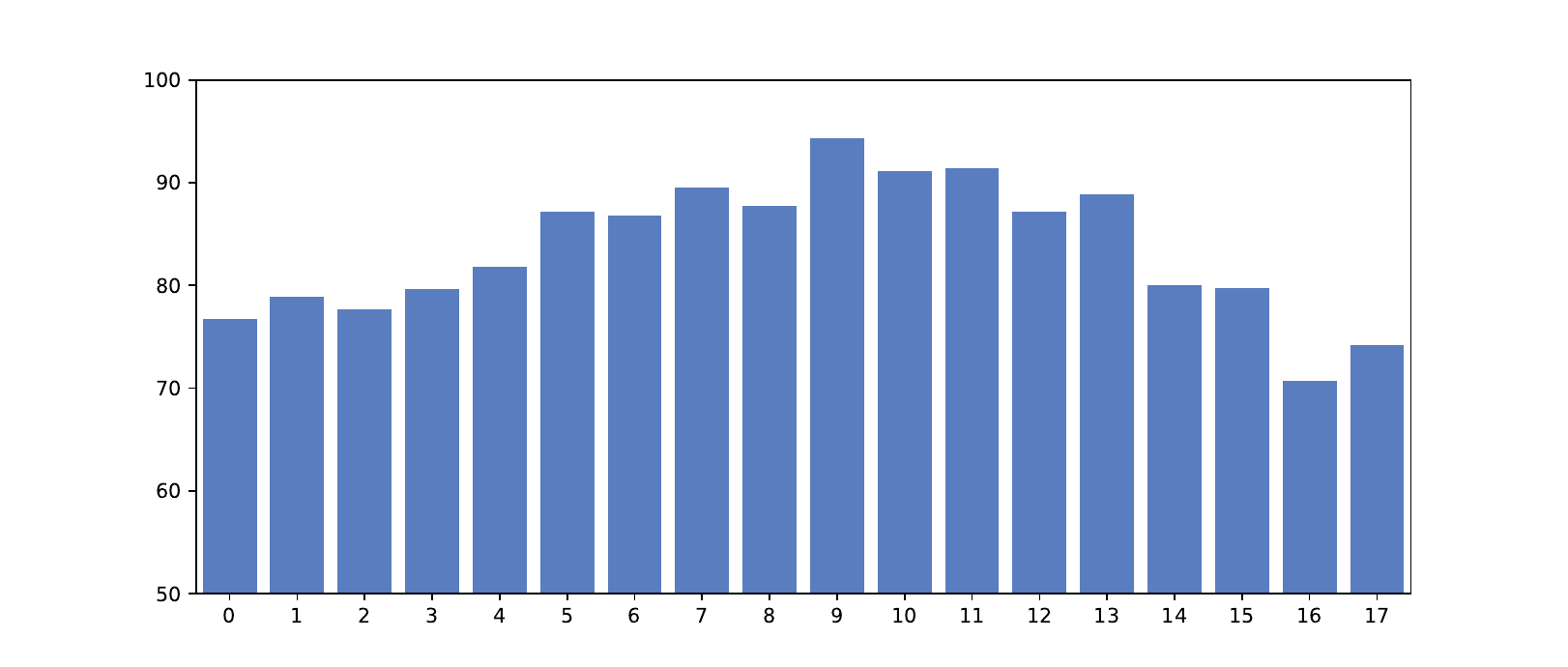}
\caption{Binary discrimination accuracy if only a single channel is selected}
\label{accvsdim}
\end{figure}

To see if further dimensionality reduction could be considered, the previous
database of latent codes of dimensionality $18\times 512$, was splitted along channels
into 18 different databases, on each of which a classifier (same architecture as the ones used before) was trained. The results of this experiment is presented Figure \ref{accvsdim}.

Reassuringly, information contained in the earlier and final channels are not the most useful when it comes to deepfake detection. Models trained on the first five or latter four channels struggle to reach 80\% accuracy. On the other hand, the tenth channel, controlling for mid-level features provides enough information to allow deepfake discriminators to reach more than 90\% accuracy. This result is sensible, as in a finished deepfake, both low and high level features of the impersonated subject are kept (global shape of the face and hair color for example) while expressions  and mid-level details are the ones being modified.

Capitalizing on newer inversion techniques and understanding the face generation process can enhance accuracy by up to 20 percentage points and reduce data dimensionality by a factor of 18.

\subsection{The Deepfake Discrimination Pipeline} \label{fullpipesubsection}
Now that the utility of projecting images in the latent space of StyleGAN is confirmed, and that a few inversion methods are found suitable for the task, a thorough comparison of the proposed method against state-of-the-art deepfake detectors is made. The benchmark focuses on the binary classification accuracy, the computational cost of the model, as well as the evolution of the performances depending on the number of training examples used.
Three state-of-the-art models are trained from scratch on twenty epochs of the CelebDFv2 dataset: XceptionNet\cite{chollet2017xception}, MesoNet\cite{afchar2018mesonet} and EfficientNet\cite{tan2019efficientnet}, as well as the proposed classifier (E2style dimensionality reduction and a Multilayer Perceptron with 5 fully-connected layers).
XceptionNet was chosen for its well known performance in deepfake detection, EfficientNet and MesoNet because of their low computing requirements. These models were trained from scratch with their respective hyperparameters left to default.
The proposed method is trained on the tenth channel of the inverted codes in W+ space (dimensionality of 512), while the CNNs are trained on images of size $256\times256\times3$.

To highlight the performances of each method on different amount of training data, discrimination accuracy has been measured on subsets of the training set. Table \ref{maintable} shows the binary discrimination accuracy on different subsets, as well as the amount of Multiply-Accumulate Operations (MACS) for each model that is trained.  Each column
represents the performance of a model trained on a specific percentage of the
training videos, ranging from a hundred to one percent of the training examples.

\begin{table}[]
\centering
\begin{adjustbox}{width=\linewidth}
\begin{tabular}{l|cccccc|c|}
  &100&50&25&10&05&01& MACS (M) \\
 \hline
MesoNet & 83.8 & 79.7 & 67.6 & - & 64.5 & 64.5 & 63 \\
EfficientNet  & 88.0 & 85.0 & 76.8 & 66.7 & 64.4 & 62.4 & 540\\
XceptionNet & \textbf{93.7} & \textbf{89.7} & 76.4 & 71.5 & 62.9 & 64.0  & 6010 \\
MLP (Ours)  & 91.5 & 87.5 & \textbf{83.4} & \textbf{77.7} & \textbf{70.0} & \textbf{65.0} & \textbf{2.79}
\end{tabular}
\end{adjustbox}
\label{maintable}
\caption{Comparison of Binary Discrimination Accuracy Across Various Data Subsets (\% of Training Examples Seen) MACS are in million of operations.}
\end{table}

It is to be noted that the proposed method requires the inversion of the database in the latent space of StyleGAN beforehand, and as such, some extra computations must be done. However, unlike the models that are compared, the inversion models are used pre-trained on real images and would not necessitate to be re-trained every time a new deepfake manipulation method emerges.
While state of the art methods achieve the best results on the full training set, their performances tend to fall when fewer training examples are available. The proposed method is competitive when all or half of the training data is available, outperforming MesoNet and EfficientNet while being behind XceptionNet by only around two percentage points. These performances are especially remarkable when the number of calculations done by the trained model are taken into account : the proposed method only computes around three million multiplications and accumulations, while XceptionNet computes around six billion of such operations. The proposed method also achieves better performance as soon as around twenty-five percent of the examples are available, and keeps outperforming other state-of-the-art models when the data is even more scarce. As such, the proposed method is better suited when a new deepfake manipulation method emerges and the amount of training data for such a manipulation is still rare. Furthermore, the lower number of operations computed by the final classifier is a necessary step towards deepfake detection democratization and ease of implementation.

Every time a new deepfake manipulation method is proposed, state of the art models need to be trained again to provide satisfactory detection results. Creating new databases is long and expensive and as such, most of the well-used databases of today were first proposed on a smaller scale (FaceForensics \cite{rossler2018faceforensics} before FaceForensics++\cite{rosslerfaceforensics}, CelebDF had a version one before being upgraded to v2, and the DFDC database \cite{dolhansky2020deepfake} was first made available through a preview subset\cite{dolhansky2019deepfake}).
To simulate such new manipulation methods, the proposed method is tested on databases of smaller scale, CelebDF v1 and DFDC preview.
The results are presented in table \ref{smalldb}

\begin{table}
\centering
\begin{tabular}{l|cc}
  &DFDC-preview&CelebDF v1 \\
 \hline
EfficientNet  & 94.42 & 85.50 \\
XceptionNet & 95.95 & 83.55 \\
MLP (Ours)  & \textbf{98.46} & \textbf{89.35}
\end{tabular}
\label{smalldb}
\caption{Comparison of Binary Discrimination Accuracies on different databases}
\end{table}
When the amount of data is limited, the proposed method consistently outperforms state-of-the-art models, by as much as six percentage points. This makes it the solution of choice when dealing with a new deepfake manipulation method, or when one is unsure which kind of deepfake is being used. Especially since the proposed method mainly uses pre-trained manipulation-agnostic models (StyleGAN and E2Style), and that the models that do need to be trained for deepfake detection require much less computational power than traditional state of the art methods.
Moreover, as shown in Table \ref{resTable}, the other recent StyleGAN inversion models still achieved better accuracy than EfficientNet and XceptionNet on this dataset, further demonstrating that the power of the proposed method lies in the data format rather than the specific model. This allows for the hope that as improvements in the domain are made (better inversion methods, or even better facial generation models), the proposed method is poised to advance accordingly.

\section{Conclusion}
We believe the proposed method can be an efficient step towards better reactivity of deepfake detection methods in the arms race against content manipulation. Lower computing requirements and the necessity of fewer training examples for comparable or even better results than state-of-the-art models may help in having an adequate response quickly to upcoming new manipulation methods. Especially since generalization is still a burning issue in deepfake detection.

\bibliographystyle{IEEEbib}
\bibliography{egbib}

\end{document}